\documentclass[runningheads]{llncs}
\usepackage{graphicx}
\usepackage{hyperref}       % hyperlinks
\usepackage{url}            % simple URL typesetting
\usepackage{booktabs}       % professional-quality tables
\usepackage{amsfonts}       % blackboard math symbols
\usepackage{nicefrac}       % compact symbols for 1/2, etc.
\usepackage{microtype}      % microtypography
\usepackage{multirow}
\usepackage{graphicx}
\usepackage{amsmath,amssymb}
\usepackage{float} 
\usepackage{subcaption}
\usepackage{multirow}
\usepackage{adjustbox}
\usepackage{hhline}  % 双线表
\usepackage{hyperref}
\usepackage{algorithm}
\usepackage{algorithmic}
\usepackage{bm}
\usepackage{bbm}
\usepackage{array}
\hypersetup{hidelinks,
	colorlinks=true,
	allcolors=black,
	pdfstartview=Fit,
	breaklinks=true}
\usepackage{diagbox}
\usepackage{color}
\usepackage[table]{xcolor}
\usepackage{wrapfig}

\newcommand{\blue}[1]{\textcolor{blue}{#1}}

\usepackage[accsupp]{axessibility}  % Improves PDF readability for those with disabilities.

\begin{document}
% \linenumbers
\pagestyle{headings}
\mainmatter
\def\ECCVSubNumber{2829}  % Insert your submission number here

\title{Exploring Resolution and Degradation Clues as Self-supervised Signal for Low Quality Object Detection} % Replace with your title

% \begin{comment}
% \titlerunning{ECCV-22 submission ID \ECCVSubNumber} 
% \authorrunning{ECCV-22 submission ID \ECCVSubNumber} 
% \author{Anonymous ECCV submission}
% \institute{Paper ID \ECCVSubNumber}
% \end{comment}
%******************

%\begin{comment}
\titlerunning{Abbreviated paper title}
\author{Ziteng Cui\inst{1} \and
Yingying Zhu\inst{2} \and
Lin Gu\inst{3,4}\thanks{Corresponding author.} \and
Guo-Jun Qi\inst{5} \and
Xiaoxiao Li\inst{6} \and \\
Renrui Zhang\inst{7} \and
Zenghui Zhang\inst{1} \and
Tatsuya Harada\inst{4,3}}
\authorrunning{F. Author et al.}
\institute{Shanghai Jiao Tong University \and
University of Texas at Arlington \and
RIKEN AIP\and The University of Tokyo \and
Laboratory for Machine Perception and Learning \and
The University of British Columbia \and
Shanghai AI Laboratory}
%\end{comment}
%******************
\maketitle

\begin{abstract}
Image restoration algorithms such as super resolution (SR) are indispensable pre-processing modules for object detection in low quality images.  Most of these algorithms assume the degradation is fixed and known a priori. However, in practical, either the real degradation or optimal up-sampling ratio rate is  unknown or differs from assumption, leading to a deteriorating performance for both  the pre-processing module and the consequent high-level task such as object detection. Here, we propose a novel self-supervised framework to detect objects in degraded low resolution images. We utilizes the downsampling degradation as a kind of transformation for self-supervised signals to explore the equivariant representation against various resolutions and other degradation conditions. The Auto Encoding Resolution in Self-supervision (AERIS) framework could further take the advantage of advanced SR architectures with an arbitrary resolution restoring decoder to reconstruct the original correspondence from the degraded input image. Both the representation learning and object detection are optimized jointly in an end-to-end training fashion. The generic AERIS framework could be implemented on various mainstream object detection architectures with different backbones. The extensive experiments show that our methods has achieved superior performance compared with existing methods when facing variant degradation situations. Code is available at \href{https://github.com/cuiziteng/ECCV_AERIS}{\blue{this link}}.

\end{abstract}

\setcounter{footnote}{0}
\section{Introduction}

\begin{figure}[!htp]
\begin{subfigure}{0.49\textwidth}
  \centering
  \includegraphics[height=5.0cm,width = 5.6cm]{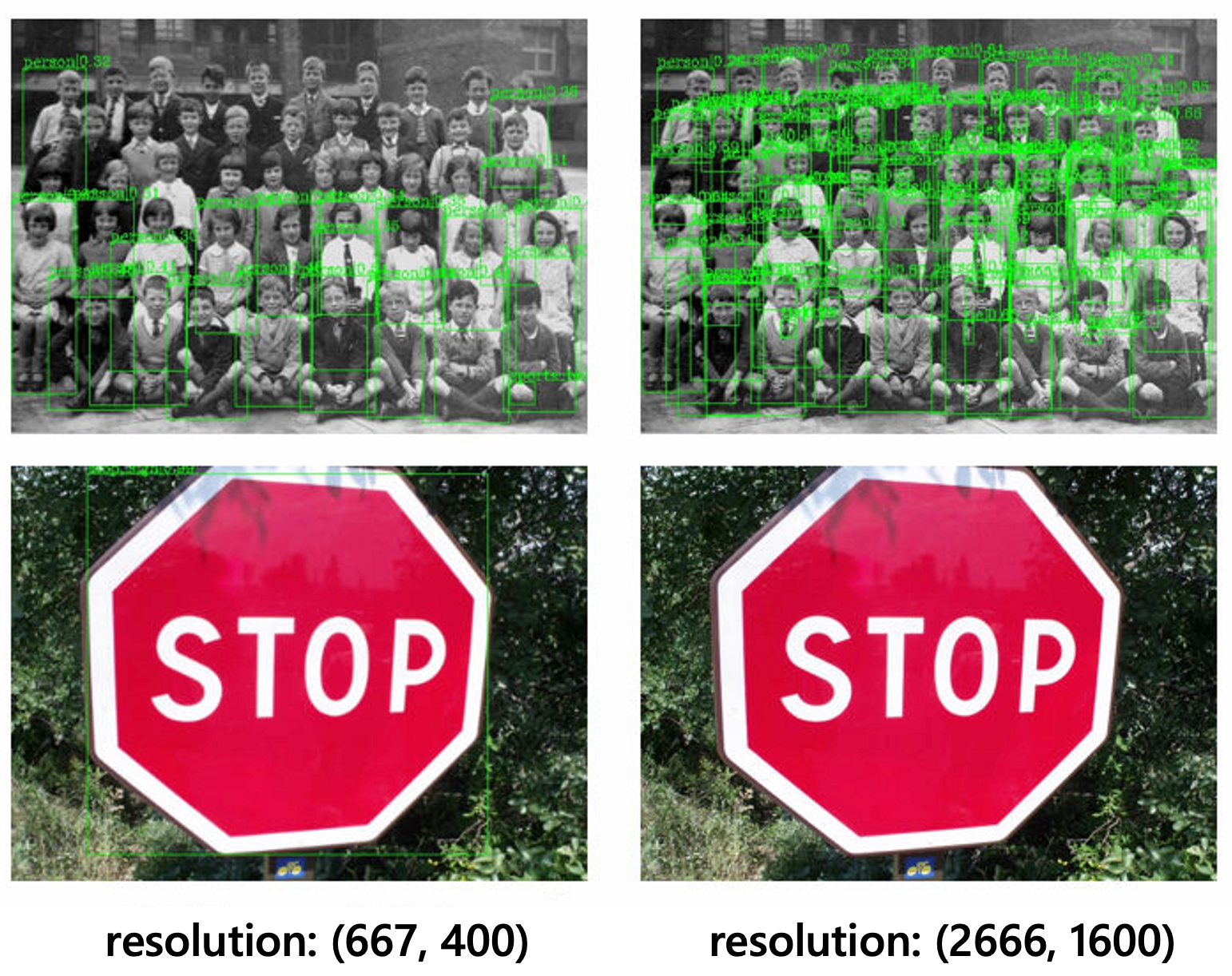}
  \caption{}
  \label{fig:moti_a}
  \end{subfigure}
\begin{subfigure}{0.49\textwidth}
  \centering
  \includegraphics[height=5.0cm,width = 5.6cm]{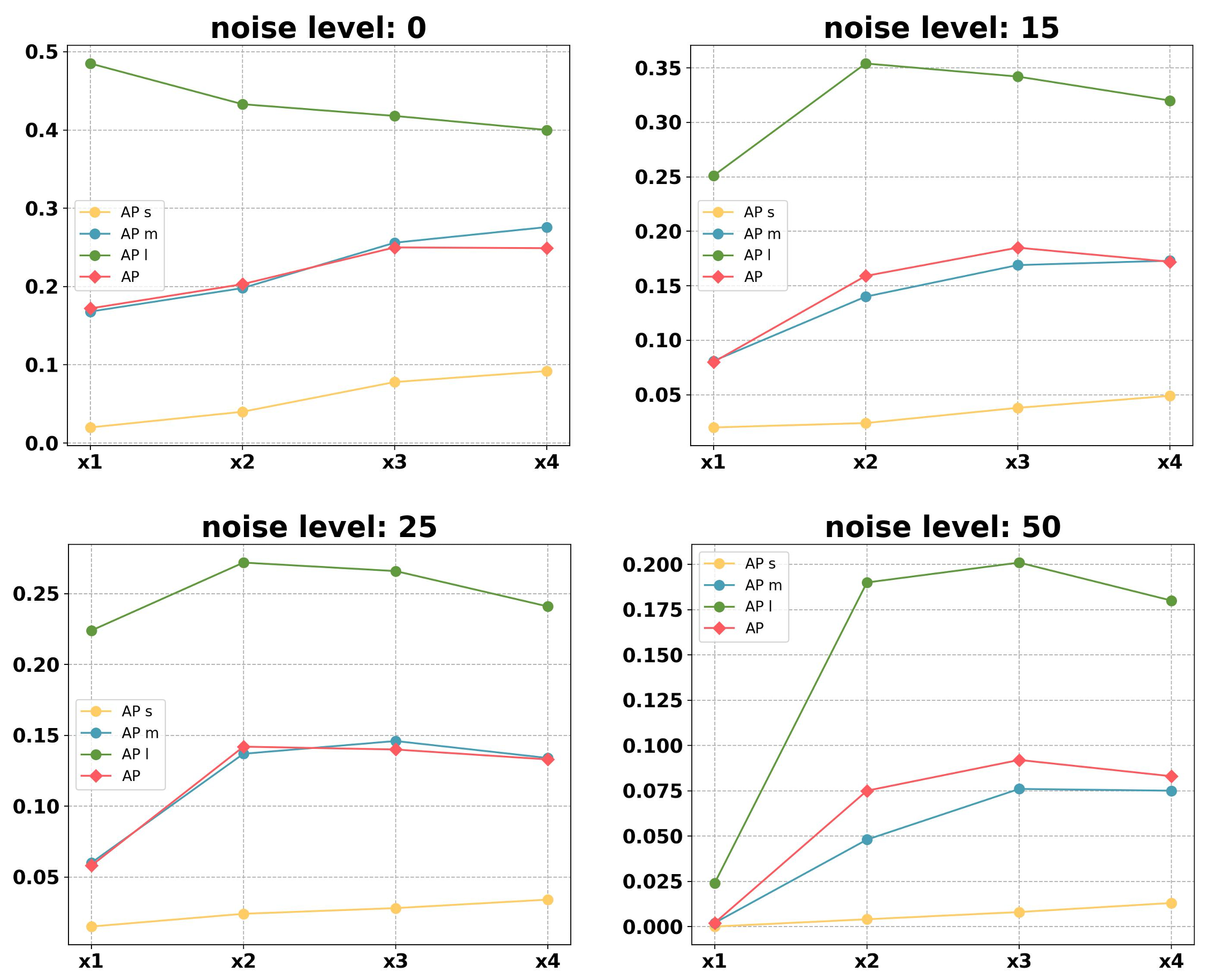}
  \caption{}
  \label{fig:moti_b}
 \end{subfigure}
 \label{fig:scale_variance}
 \caption{Illustration of scale variance bottleneck. (a): Tiny people are well detected at high resolution while the large traffic sign is recognized in low resolution. (b): Detection result on down-sampled MS COCO~\cite{coco_dataset} dataset (down scale rate: 4) with different noise level. Specifically, we up-scale the images with different ratio (2, 3, 4) before detection. X-axis is up-scale ratio and Y-axis is mAP result of CenterNet~\cite{centernet}. We also report the results on small, medium and large objects.}
\end{figure}

High level vision tasks (\textit{i.e.} image classification, object detection, and semantic segmentation) have witnessed great success thanks to the large scale dataset \cite{imagenet,coco_dataset,voc_dataset}. Images in these datasets are  mainly captured by commercial cameras with higher resolution and signal-to-noise ratio (SNR). Trained and optimized on these high-quality images, high-level vision would suffer a performance drop on low resolution ~\cite{SR_for_vision_tasks,SR_object_detection,SR_for_segmentation} or low quality images~\cite{noise_4_vision,Blur_4_vision,CVPR_blur_detection,White_balance,ICCV_MAET,Ma_2022_CVPR}. 

To improve the performance of vision algorithms on degraded low resolution images,  Dai \textit{et al.}~\cite{SR_for_vision_tasks} presented the first  comprehensive study advocating pre-processing  images with super resolution (SR) algorithms. Other high-level tasks like face recognition~\cite{low_face_recognition}, face detection~\cite{Low_face_CVPR19}, image classification~\cite{low_image_classification,noise_4_vision} and semantic segmentation~\cite{SR_for_segmentation}, also benefit from the restoration module to extract more discriminate features. 

Most existing enhancement methods, especially SR algorithms~\cite{SR_kernelGAN,SR_USRNet,SR_cvpr2021_random}, assume target  images are from a \textbf{known and fixed} degradation model~\cite{Degradation_model,Degradation_model_1}: 
\begin{equation}
    t(x) = (x \circledast k)\downarrow_s + n,
    \label{eq:degradation_model}
\end{equation}
where $t(x)$ and $x$  denote the degraded low resolution (LR) image and original high resolution (HR) input respectively. $k$ is the blur kernel while $\downarrow_s$ is the down-sampling operation with ratio $s$.  $n$ is the additive noise. However, the performance of these enhancement algorithms would decline severely when the real degradation deviates from the assumption~\cite{SR_Iter_Kernel}. To make it worse, for machine perception tasks, as shown in Fig~\ref{fig:moti_b}, higher resolution does not necessarily guarantee a better performance in high level tasks. Like object detection, the optimal SR ratio varies across the images due to the scale variance bottleneck~\cite{SNIP,SNIPER}, there is a trade off that certain high level predictions are better handled at lower resolution and others better processed at higher resolution. As illustrated in Fig~\ref{fig:moti_a}, though working well on individual tiny person at high resolution (2666,1600), the detection method ignores the large traffic sign. On the contrary, in low resolution (667,400) images, detecting network's reception field could observe more global context for large structure, at the cost of sacrificing the  small objects. Fig~\ref{fig:moti_b} also quantitatively demonstrates this bottleneck. The detection performance does not necessary increases with super resolution ratio, especially for large objects.

%To make it worse, since these restoration methods are human visual perception oriented, they are often inadequate for machine perception tasks such as object detection~\cite{ICCV_MAET,TIP20_lowvisibility,resizer_2021_ICCV}.

\begin{figure}
    \centering
    \includegraphics[width = 8.9cm]{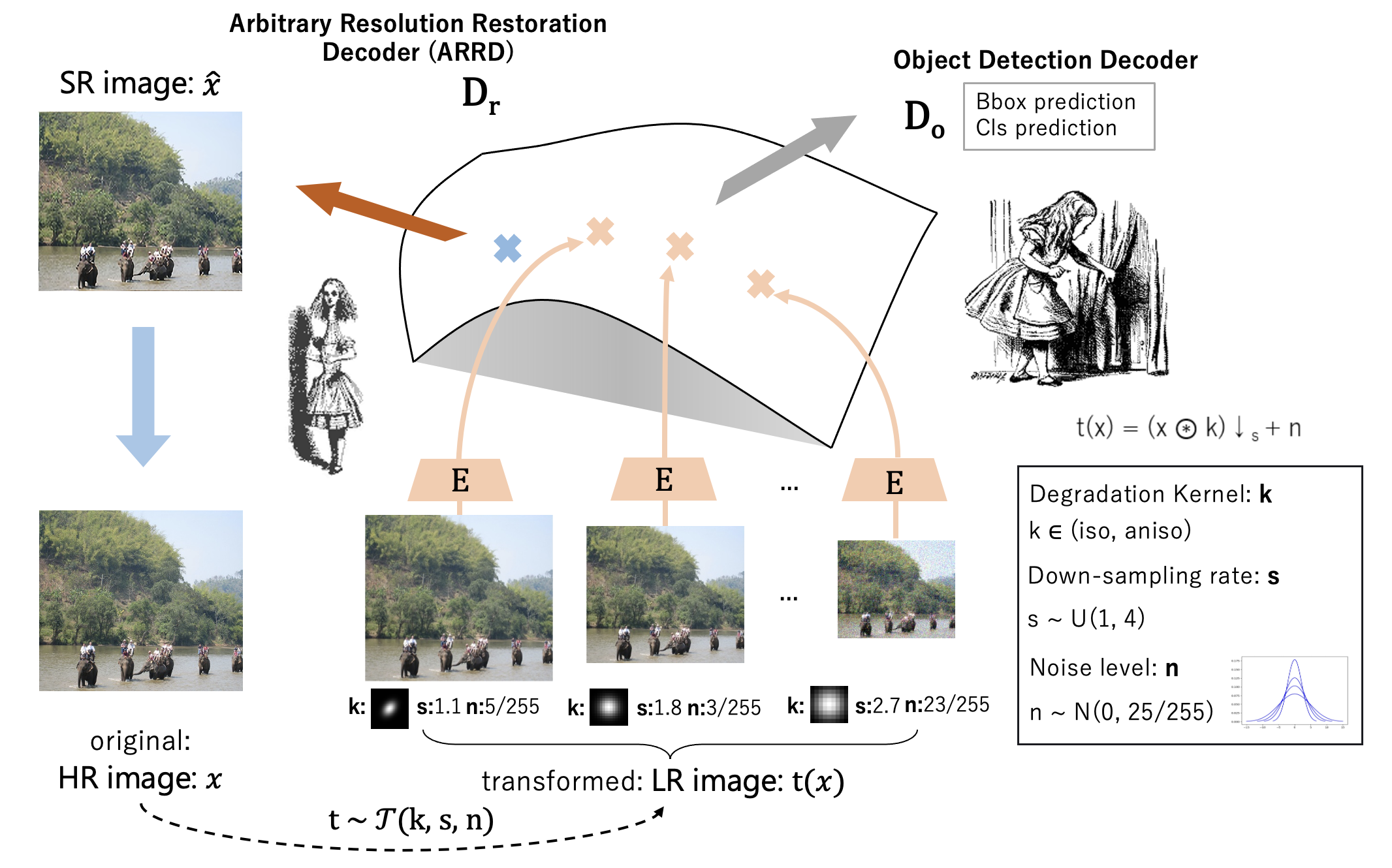}
    \caption{A simple illustration of Auto Encoding Resolution in Self-supervision (AERIS). Encoder $E$ encodes images transformed under various self-supervised signal into degradation equivariant representations (like Alice after drinking ""Drink me""). ARRD $D_r$ implicitly decodes the self-supervised signal to reconstruct the original image/transformation while detection decoder $D_o$ decodes the representations for object detection task.}
    \label{fig:repre_manifold}
\end{figure}

Instead of explicitly enhancing an input image with a fixed restoration module, we exploit the intrinsic equivariant representation against various resolutions and degradation. \textit{I know who I was when I got up this morning, but I think I must have been changed several times since then.}\footnote{Chapter 5, Alice in Wonderland}. Either being small enough to squeeze through the door or so big to shed a pool of tears, Alice
should be encoded with a equivariant representation to show who she is in the world. Based on the encoded representation shown in Fig.\ref{fig:repre_manifold}, we propose an end-to-end framework for object detection in low quality images. To capture the complex patterns of visual structures, we utilize groups of downsampling degradation transformations under different downsampling rate, noise and degradation kernel as the self-supervised signal~\cite{aet,RotNet}.

During the training, we generate a degraded LR image $t(x)$ from the original HR image $x$ through a random degradation transformation $t$. As shown in Fig.\ref{fig:repre_manifold}, to train the Encoder $E$ to learn the degradation equivariant representation  $E(t(x))$, we introduce an arbitrary-resolution restoration decoder (ARRD)  decoder $D_r$. ARRD implicitly decodes $t$ to reconstruct the original HR data $x$ from the representation $E(t(x))$ of various degraded LR image $t(x)$. If the self-supervised signal is reconstructed, the representation should capture the dynamics of how they change under different resolution and other degradation as much as possible~\cite{group_convolution,capsule_network,aet}. The nature of reconstructing HR data also allows us to leverage the advance of the fast-growing SR research by directly using their successful architectures. 

On the encoded representation $E(t(x))$, we further impose an object  detection decoder $D_o$  to supervise the encoder $E$ to encode the image structure relevant to the consequent tasks. The object  detection decoder $D_o$ performs the detection task to get the object's location and class. During inference, the target image is directly passed through the encoder $E$ and object detection decoder $D_o$ in Fig.\ref{fig:repre_manifold} for detection. Compared to pre-processing module based methods~\cite{Aerial_detection_SR,SR_object_detection},  our inference pipeline is more computation efficient as we avoid explicitly reconstructing the image details.

%ARRD $D_r$ would also supervise the encoder $E$ to encode the detailed image structure which facilitates the consequent tasks. In addition, 

%to decode both the  applied degraded transformation $t$ 

%decode the applied degraded transformation $t$ from the representation $E(x)$ and $E(t(x))$. If the transformation could be reconstructed, the representation should capture the dynamics of how they change under different transformations as much as possible~\cite{group_convolution,capsule_network,aet}.

%To further take advantage of the strength of the fast-growing SR research, we introduce an arbitrary-resolution restoration decoder (ARRD) $D_r$ Fig.\ref{fig:repre_manifold}.

%ARRD reconstructs the original HR data $x$ from the representation $E(t(x))$ of various degradated LR image $t(x)$. ARRD $D_r$ would supervise the encoder $E$ to encode the detailed image structure which facilitates the consequent tasks.

To cover the diverse degradation and  resolutions, in real scenario, we generate degraded $t(x)$ by randomly  sampling a transformations $t$ according to practical down-sampling degradation  model~\cite{Degradation_model_1,SR_cvpr2021_random}. As shown in Fig.\ref{fig:repre_manifold},  the transformation $t$ is characterised by down-sampling ratio $s$, blur kernel $k$,  and noise level $n$ in Eq.\ref{eq:degradation_model}. 

Our contributions could be summarised as follows:

\begin{itemize}
    \item We propose a novel framework, Auto Encoding Resolution in Self-supervision (AERIS), to detect objects in degraded low resolution images by utilizing the resolution and degradation clues as self-supervised signal. Specifically, we learn the degradation equivariant representation that captures the dynamics of feature representations under  diverse resolutions and degradation types. Our AERIS is generic and could be implemented on several mainstream object detection architecture.
   \item AERIS method takes the strength of advanced super resolution (SR) research by training an arbitrary resolution restoration decoder (ARRD) that reconstructs the high resolution details. Furthermore, by optimizing the representation learning and detection in a unified end-to-end training framework, the representation preserves the intrinsic visual structure that is discriminative for detection.
    \item   We evaluate our method on mainstream public dataset MS COCO~\cite{coco_dataset} and KITTI~\cite{kitti_dataset}. The experiment results also show that our method has achieved SOTA performance on several degradation conditions.
\end{itemize}

\section{Related Works}

\subsection{Single Image Super Resolution}

%% Basic SR

Image restoration algorithms are intuitive solutions to handle degradation, here we mainly introduce single image super-resolution (SISR), since the object detection task is sensitive to resolution, and the other low-level vision tasks (\textit{i.e.} denoise, deblur) also have connections with SISR task. 

The very first CCN-based SR was proposed by Dong \textit{et al.}~\cite{SRCNN} with a three-layer neural network. Then Kim \textit{et al.}~\cite{SR_deepCNN} extended the depth of network to 20 layers with gradient clipping  and residual learning. Batch normalization is later identified to impose the negative effect on the SR reconstruction. By removing this layer, EDSR~\cite{SR_EDSR} achieves SOTA in 2017. After ESDR, better SR architectures are designed by integrating the successful deep learning techniques such as Laplacian pyramid structure~\cite{SR_Laplacian}, dense connection~\cite{SR_densenet}, back projection~\cite{SR_DBPN}, transformer blocks~\cite{SR_SWIN-IR} and so on. Besides designing sophisticated architecture, losses like perceptual loss~\cite{SR_perpectual_loss} and adversarial loss~\cite{SRGan}  are also demonstrated to improve the SR reconstruction quality. 

%% Arbitrary Resolution Super-resolution
SR algorithms heavily rely on the assumption of degradation model and fixed resolution. Much efforts are spent to relax the constraint. Not limited to a unfixed up-sampling scale, like Hu \textit{et al.}~\cite{SR_arbitary_MetaSR} first proposed Meta-SR to super-resolve images with arbitrary scale factor. After that, LIIF~\cite{SR_arbitary_LIIF} utilize implicit function to solve this task, and FuncNet~\cite{FuncNet} further been proposed also for noise and blur condition. 

%% Real-World SR
To deploy SR for real scenarios, blind SR methods assume the degradation information is not known. One direction is to convert the problem into non-blind SR by provide prior degradation information~\cite{SR_SRMD} or initially estimate the degradation parameters~\cite{SR_kernelGAN}. However, the applied non-blind SR algorithm is very sensitive to the error of the degradation estimation.  Gu \textit{et al.}~\cite{SR_Iter_Kernel} then proposed to iterative correct the estimated degradation with an iterative kernel correction (IKC) method. Without explicitly estimating degradation parameters,  Wang \textit{et al.}~\cite{Wang_2021_CVPR_contrastiveSR} introduced a contrastive loss to design the degradation-aware SR network based on the learned representations. Recently, Zhang \textit{et al.}~\cite{SR_cvpr2021_random} solved the general blind SISR by designing a practical model considering complex degradation. This model has been demonstrated to cover the degradation space of real images. Therefore, we adopt this practical model to synthesize various degraded LR images as the self-supervised signal to train our model.

\subsection{Image Restoration for Machine Perception}

There is sufficient evidence that degraded scene would give negative impact on high-level vision tasks~\cite{TIP20_lowvisibility,SR_for_vision_tasks,SR_object_detection,robust_segmentation_1}. As for resolution, Dai \textit{et al.}~\cite{SR_for_vision_tasks} made the first analyze on improving several vision tasks with SR as pre-process. Wang \textit{et al.}~\cite{low_image_classification} analyzed the effectiveness of SR in image classification task while DSRL~\cite{SR_for_segmentation} improved the low-resolution semantic segmentation with an additional SR block. Shermeyer and Etten~\cite{Aerial_detection_SR}  evaluate the effectiveness of a SR pre-process step on aerial image object detection. Recently, Haris \textit{et al.}~\cite{SR_object_detection} jointly optimise object detection loss along with SR sub-network~\cite{SR_DBPN} to improve detection performance.

Similarly, noise and blur's effect on high-level vision have also been well-studied. Hendrycks \textit{et al.}~\cite{ICLR_2019_robustness} evaluate image classification robustness under multiply degradation conditions including noise and blur. Kamann \textit{et al.}~\cite{robust_segmentation_1} studied the impact of noise and blur on different semantic segmentation methods.  Liu \textit{et al.}~\cite{noise_4_vision} combine a denoise network in classifier to improve classification's performance under noisy condition.  Very recently, Mohamed and Gabriel~\cite{CVPR_blur_detection} analyse  motion blur and propose several methods to improve detection performance on motion blurry images.

However, most of these existing works assume the degradation parameters such as the down-sampling ratio is known and fixed. Based on the degradation equivariant representation, our framework is robust to various degradation in real-world scenarios. Without an explicit restoration module or restoration step, we directly perform the detection on low-dimension encoded features that saves much computational burden.

\section{Down-sampling Degradation Transformations}
\label{sec:degradation}

\begin{figure}
    \centering
    \includegraphics[width=10cm, height=5cm]{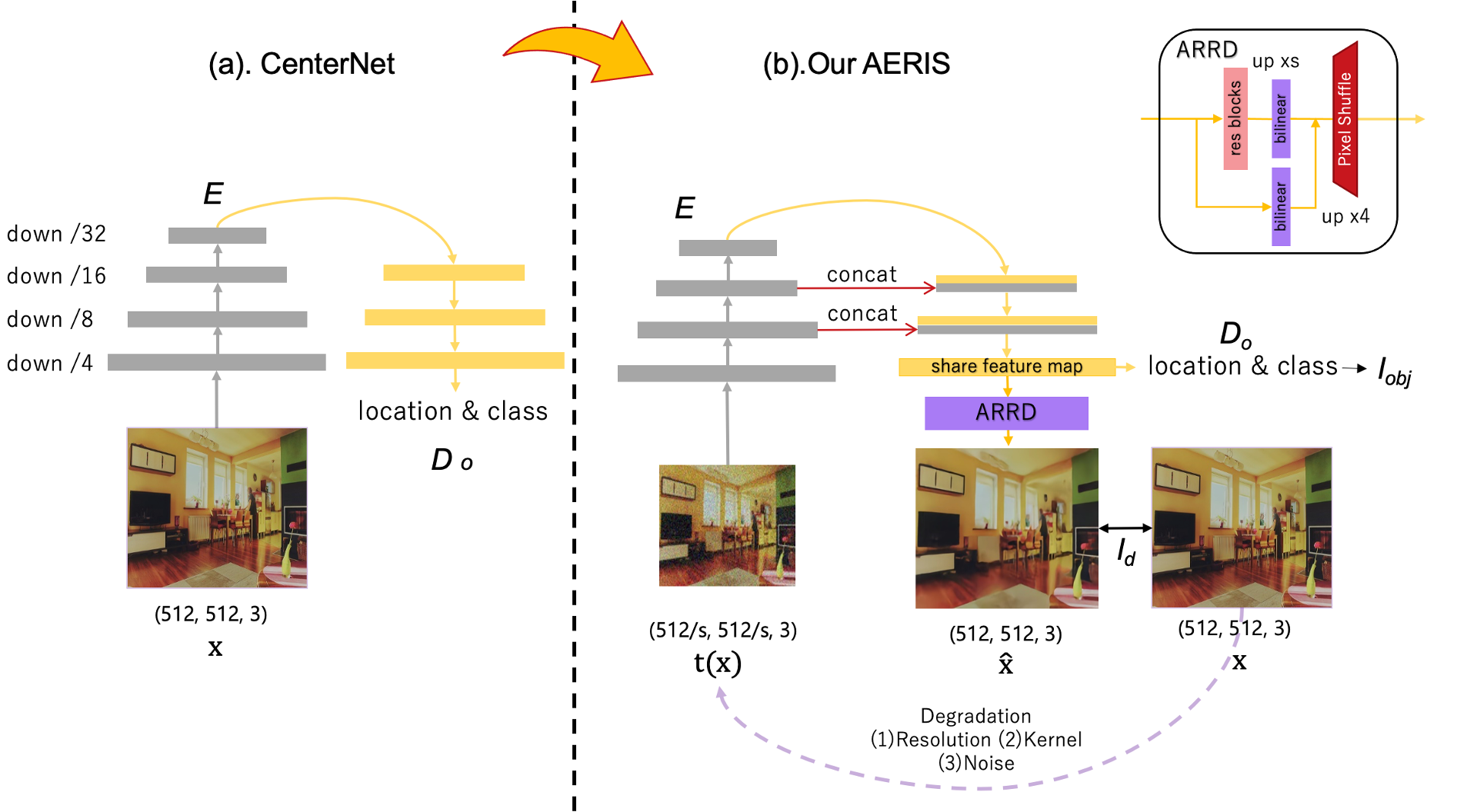}
    \caption{An illustration of how our AERIS is implemented based on CenterNet~\cite{centernet}. Left is the original CenterNet while the right one is the architecture of AERIS.}
    \label{fig:detail_structure}
\end{figure}

In real scenarios, the image may be captured and processed in various ways. To cover these generally unknown operations, it is necessary to select a practical degradation model for the degradation transformation. This model would  transform the high resolution (HR) image $x$ to the low resolution counterpart $t(x)$ with Eq.~\ref{eq:degradation_model}.

Early restoration methods assume a simple degradation model where LR is directly down-sampled from the HR images without or with simple noise. Instead of dealing with synthetic images, recent methods now focus on more realistic degradation models. For example, \cite{SR_real_images_1} directly train the model on the  LR and HR images pair captured by the real camera system.  USRNet~\cite{SR_USRNet} effectively handled the degradation models with different parameters such as scale factors by unfolding the model-based energy function. 
Here, we adopt the practical degradation model~\cite{Degradation_model,Degradation_model_1,SR_kernelGAN,SR_USRNet,SR_cvpr2021_random} that accounts for diverse degradation in real images.

\textbf{Convolution Operation:} Gaussian blur kernel is the most common kernel to blur the image~\cite{gaussian_kernel,gaussian_kernel_2,SR_kernelGAN}. Here, we choose two Gaussian degradation kernels $k$: isotropic Gaussian kernels $k_{iso}$ and anisotropic Gaussian kernels $k_{aniso}$~\cite{SR_kernelGAN,SR_SRMD,SR_cvpr2021_random}. We also consider none degradation kernel as $k_{none}$. Following ~\cite{SR_cvpr2021_random}, the kernel size is uniformly sampled from $\left\{7\times7, 9\times9, ..., 21\times21 \right\}$ and the $k_{iso}$'s width is uniformly chosen from $(0.1, 2.4)$. For $k_{aniso}$, the kernel angle is uniformly chosen from $(0, \pi)$ and the longer kernel width is uniformly chosen from $(0.5, 6)$.

\textbf{Noise:} When dealing with the real-world scenarios,  the Gaussian noise model is usually adopted to simulate the noises from  camera sensor noise~\cite{camera_sensor_noise}, low-light noise ~\cite{low_light_noise} to quantization noise ~\cite{quantisation_noise} \textit{etc.} Therefore, we adopt a zero-mean additive white Gaussian noise (AWGN) model $n \sim N(0, \sigma)$ in Eq.\ref{eq:degradation_model}. The variance $\sigma$ is randomly chosen from a uniform distribution $U(0, 25/255)$ (\textit{e.g.} $13.2/255$). 

% Here we take the noise level $n$ as signal to learn variance $\sigma$ for self-supervised representation, the variance $\sigma$ has been normalized in $(0,1)$ to represent noise level.

\textbf{Down-sampling:} For the down-sampling process, the sampling ratio $s$ is randomly chosen from uniform distribution $U(1, 4)$ (\textit{e.g.} 1.9) while the down-sampling  methods is randomly chosen from nearest method $d_{nearest}^s$, bilinear method $d_{bilinear}^s$ and bicubic method $d_{bicubic}^s$.

% Here we take the down-sampling ratio $s$ as the self-supervised signal to learn its reciprocal $\frac{1}{s}$, it worth noting that the the down-sampling ratio $s$ per batch should keep same for network forward. 

So the final down-sampling degradation transformation $t$ would take from random blur kernel $k$, noise level $n$ and down-sampling ratio $s$ in the total distribution $\mathcal{T}$, as $t \sim \mathcal{T}(k,n,s)$.

\section{Our Framework}

Due to its concise structure, we mainly take one-stage object detector CenterNet~\cite{centernet} to illustrate how to implement our AERIS. More object detectors' implements and results please refer to supplementary. We think that our AERIS framework is a generic framework that could also be implemented on other popular object detectors~\cite{detection_yolo,detection_fast_rcnn} and various backbones~\cite{mobile_net,Swin_Transformer,li2022uniformer,mao2021dualstream}.

%Here, we show how to implement our framework in an encoder-decoder style structure.

\subsection{CenterNet}

CenterNet~\cite{centernet} is an efficient one-stage anchor-free object detector. We show its vanilla  structure of~\cite{centernet} in Fig.\ref{fig:detail_structure} (a). Input image is fed to the backbone (\textit{i.e.}  ResNet18~\cite{resnet}) to extract $/32$ bottleneck feature, and then  upsampled to a $/4$ feature map by three $\times2$ deconvolution blocks.  This $/4$ feature map is  passed to scaling blocks with three independent convolution blocks to generate  the final feature maps. Based on this feature map,  there are prediction heads conducting class-wise bbox center detection, bbox height and width regression, offset regression respectively. For more details, please refer to ~\cite{centernet}.

The CenterNet could be decomposed into an encoder-decoder style structure. Here, we denote the network backbone part (gray part in Fig.\ref{fig:detail_structure}.a) as encoder $E$. The object detection decoder $D_o$, comprised by three prediction heads (colored in orange in Fig.\ref{fig:detail_structure} (a)), decodes the object information.

\subsection{Architecture and Training Pipeline}

 Fig.\ref{fig:detail_structure} (b) illustrates how to implement our AERIS based on CenterNet~\cite{centernet}. The detailed training procedure is given in Algo.\ref{RestoreDet_Algorithm}. When training AERIS, we first apply the degradation transformation $t \sim \mathcal{T}(k,n,s)$ convert $x$ to a random generated $t(x)$, covering the degradation space of real scenarios. In each batch $B$, the generated $t(x)\in \mathbb{R}^{B \times 3 \times \frac{H}{s} \times \frac{W}{s}}$ should keep the same down-sampling rate $s \sim (1,4)$. Then the transformed degraded LR image $t(x)$ are sent to the encoder $E$ to encode the degradation equivariant representation $E(t(x)) \in \mathbb{R}^{B \times C \times \frac{H}{s*d} \times \frac{W}{s*d}}$. Here the encoder $E$  refers to the backbone in detection network, and $d$ is the down-sample rate of backbone (\textit{i.e.} 32 for ResNet~\cite{resnet}). 
 
%  Here, we directly use the encoder $E$ of CenterNet but duplicate it into a shared weight Siamese structure to receive the  HR and LR image, respectively. 

%  With this encoder $E$, the encoded representative $E(x)$ and $E(t(x))$ would have same channel but different size. Then we apply a global average pooling layer $G$ to pool them into the same shape 2-D tensor.   $k$ represents the Gaussian kernel size, $s$ represents the down-sample factor and $n$ represents noise level. While training, all the ground truth $k$, $n$, $s$ is normalized to $0 \sim 1$ in their own type. %For transformation decoder $D_t$, we have  MSE loss $l_{trans}$ between $(\hat{k}, \hat{s}, \hat{n})$ and $(k, s, n)$.

We then input degradation equivariant representation $E(t(x))$ into upscaling blocks with three $\times2$ deconvolution blocks to generate the final feature map for object detection decoder $D_o$ and arbitrary resolution restoration decoder (ARRD) $D_r$. As illustrated by the red arrows in Fig.\ref{fig:detail_structure}, we introduce the skip connection on  $/8$ and $/16$ feature maps between the backbone encoder $E$ and the deconvolution blocks. Fusing  features  from different scales is a common process in low-level vision tasks, and could enhance semantic information and contribute to the subsequent $D_o$ and $D_r$.

We further regularise the representation encode $E$ with our unique ARRD $D_r$ that implicitly estimate the self-supervised singal $t$ to recover HR image $\hat{x}$. Since the downsampling rate $s$ is not a fixed integral number in the training stage, ARRD could deal with an arbitrary scale factor. ARRD $D_r$ could force the encoder $E$ to not only  capture the dynamics of how images  change under different transformations, but also extracts the  complex patterns of visual structures. Since ARRD $D_r$ aims to recover  the original resolution of clean image $x$ from $E(t(x)))$, it could also  support the object detection decoder $D_o$ with more detailed features. Inspired by the learnable resizer model~\cite{resizer_2021_ICCV}, we design this decoder with a residual bilinear model shown in Fig.~\ref{fig:detail_structure}, which ends up with a $\times 4$ pixel shuffle layer~\cite{SR_pixel_shuffle}. ARRD is a light weight structure that uses fewer parameters (0.06M) compared to the backbone encoder (11.17M), upscaling blocks (3.61M) and detection decoder $D_o$ (0.12M). %The light weight could  we more want ARRD to help $D_o$ to improve machine vision performance rather than generate images to match people's visual experience.
The ARRD loss $l_{d}$ is defined as an L1 loss between output image $\hat{x}$ and ground truth image $x$:

\begin{equation}
\begin{aligned}
    l_{d} &= |\hat{x} - x|_1 = |D_r(E(t(x))) - x|_1.
\end{aligned}
%\label{eq:loss_function}
\end{equation}

% \\
%     l_{trans} &= ||D_{t}\left[ E(t(x)), E(x)\right] - (k, s, n)||_2^2

We adopt the three CenterNet prediction heads as the object detection decoder $D_o$ to conduct detection on the final feature map generated by the upscaling block.

As shown in Algo.\ref{RestoreDet_Algorithm}, we optimise the total loss $l_{total}$ including detection loss $l_{obj}$ (\textit{i.e.} classwise bbox center loss, bbox width and height loss, bbox offset loss for CenterNet~\cite{centernet}),  and data restoration loss $l_{d}$:
\begin{equation}
    l_{total} = l_{obj}  + \lambda \cdot l_{d},
\end{equation}
where $\lambda$ is the non-negative parameters for loss balancing. Which we set to 0.4 in CenterNet experiments and 0.8 in DETR experiments, more ablation details please refer to supplementary.

\subsection{Inference Procedure}

The inference procedure only involves  encoder $E$ , upscaling block and  object detection decoder $D_o$ as illustrated in Fig.\ref{fig:detail_structure}. Specifically, the encoder $E$ encodes the input target image before $D_o$ performs the detection. Compared to explicitly pre-processing image for high-level tasks~\cite{SR_for_vision_tasks,Aerial_detection_SR,SR_object_detection}, our AERIS saves much computational time as we avoid reconstructing HR details of data.

%\textcolor{red}{Say something about saving the time and computation.}

%the target image $t(x)$ is send to the ResNet backbone encoder $E$ to get latent feature $E(t(x))$.   Object detection decoder $D_o$ then outputs the detection results (bbox location and class).  

\begin{figure}[t]
    \centering
    \includegraphics[width=1.0\linewidth]{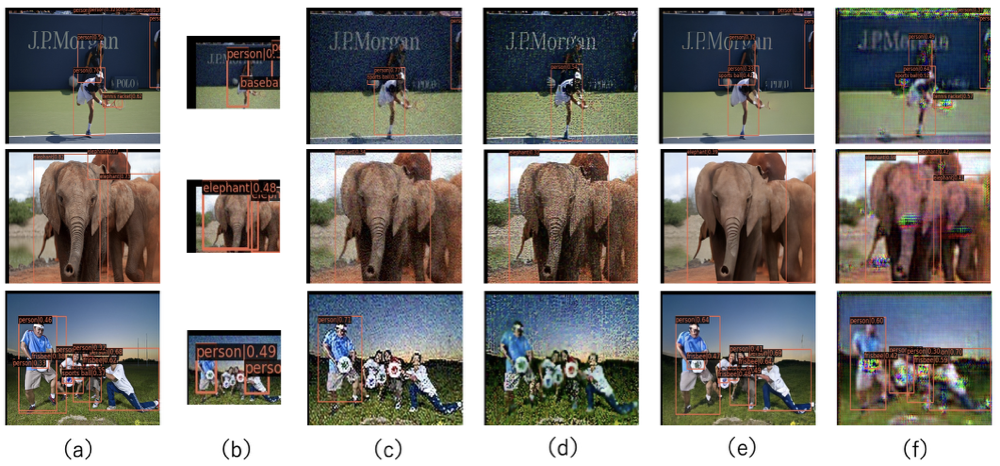}
    \caption{Exemplar detection results on MS COCO 2017 dataset~\cite{coco_dataset}. (a)/(b) is CenterNet trained on normal images and tested on normal/\textbf{COCO-d} dataset, (c)/(d)/(e) is CenterNet tested on the degraded image restored by individual SR algorithm SRGAN~\cite{SRGan}/Real-SR~\cite{SR_real_images_1}/BSRGan~\cite{SR_cvpr2021_random}. (f) is the detection result of our AERIS and we use the output of ARRD $D_r$ as background images.}
    \label{fig:results}
\end{figure}

We could also reconstruct the HR image with our ARRD decoder $D_r$.  Very interestingly, our restored images $\hat{x}$ are more machine vision oriented and exhibit artifacts around the center of the object, as shown in Fig.\ref{fig:results}.

\begin{algorithm}[t]
\caption{AERIS Algorithm Pipeline}
\label{RestoreDet_Algorithm}
\textbf{(1). Data Generation}: \\
\small B: batch size, C: channel, H: image height, W: image width\\
\normalsize \textbf{inputs:} HR image $x=(B, C, H, W)$, down-sample factor $s \sim (1.0, 4.0)$ \\
\textbf{outputs:} degraded LR image $t(x)=(B, C, \frac{H}{s},\frac{W}{s})$ 
\begin{algorithmic}
\FOR{$i$ in range($B$):}
\STATE (1). Convolution with blur kernel $k$
\STATE (2). Down-sampling with rate $s$
\STATE (3). Add noise $n$
\ENDFOR\\
\end{algorithmic}
\textbf{(2). Training}:\\
\textbf{inputs:} Degraded LR image $t(x)=(B, C, \frac{H}{s}, \frac{W}{s})$ \\
\textbf{outputs:} detection output, estimated SR image $\hat{x}$\\
\textbf{encoding:} \\
$ t(x) \underrightarrow{\quad E \quad} E(t(x)) \\$
\textbf{decoding:} \\
data restoration decoding: $\hat{x} = D_r(E(t(x))$ \\
detection decoding: detection results $= D_o(E(t(x))$
\end{algorithm}

%transformation decoding: $\hat{t} = D_t([E(x), E(t(x))])$ \\

\section{Experiments and Details}
\subsection{Datasets and Implementation Details}
\label{sec:train_details}

\textbf{Dataset.} We adopt two widely used object detection datasets MS COCO~\cite{coco_dataset} and KITTI~\cite{kitti_dataset} for detection robustness evaluation. MS COCO contains $\sim$118k images with bounding box annotation in 80 categories. We use COCO~\texttt{train2017} set as train set and use COCO~\texttt{val2017} set as normal condition evaluation set. Also COCO dataset differentiates the labels of different scale level's objects, and give them specific evaluation metrics (small: AP$_\text{s}$, middle: AP$_\text{m}$, large: AP$_\text{l}$), which could show us different degradation conditions' influence on different scale objects, especially the down-sampling process $s$.

KITTI~\cite{kitti_dataset} is a popular small object detection dataset for autonomous driving. For KITTI dataset, we evaluate car class, and use KITTI~\texttt{train} set as train set and use KITTI~\texttt{val} set as normal condition evaluation set, and show the AP rate for comparison.

\textbf{Impalement Details.}
We build our framework based on the open-source object detection toolbox \texttt{mmdetection}~\cite{mmdetection}. Throughout the experiments, the backbone ResNet-18~\cite{resnet} and Swin-T~\cite{Swin_Transformer} are initialed with ImageNet~\cite{resnet} pre-train weights. We apply the data augmentation pipeline in \texttt{mmdetection}~\cite{mmdetection}, specifically we adopt random crop, random flip and multi-scale test.

During training stage, all the models are trained on 4 Tesla V100 GPUs. Same as setting in~\cite{mmdetection}, for AERIS-CenterNet training, the input image shape is resized to 512$\times$512. The model has been trained for 140 epochs with SGD optimizer. Batch size is set to 16 per GPU. Momentum and weight decay are set to 0.9 and 1e-4. Initial learning rate is 0.01 and warms up at first 500 iterations and would decays to one-tenth at 90 and 120 epoch. 

\begin{table}[t]
\renewcommand\arraystretch{1.43}
\centering
\caption{Comparison with SOTA restoration methods and different training strategies on \textbf{COCO-d} dataset. Here \textbf{CenterNet} with \textbf{ResNet-18} backbone and \textbf{Swin-T} backbone were adopted. The inference images with \textbf{higher resolution} are in \textbf{blue} background.}
\label{tab:COCO-d}
\resizebox{\linewidth}{!}{%
\begin{tabular}{c|c|c|cccc|cccc}
\toprule
\multirow{2}{*}{Test Set} & \multirow{2}{*}{Pre-process} & \multirow{2}{*}{Training Strategy} & \multicolumn{4}{c|}{CenterNet (ResNet-18)} & \multicolumn{4}{c}{CenterNet (Swin-T)}                                                      \\ \cline{4-11} 
                          &                                     &    & AP &  AP$_\text{s}$ & AP$_\text{m}$   & AP$_\text{l}$  &  AP &  AP$_\text{s}$        &  AP$_\text{m}$    & AP$_\text{l}$ \\ \hline
COCO                      & \multirow{2}{*}{-}   & \multirow{10}{*}{Detection}        & 30.1      & 10.6     & 33.2     & 47.2     & 36.9                & 17.9                 & 41.8                 & 52.9                 \\ \cline{1-1} \cline{4-11} 
\multirow{13}{*}{\textbf{COCO-d}}  &                                     &                                    & 14.5      & 1.2      & 10.4     & 38.6 & 19.9 &  2.7  &  16.9  &  46.2  \\ \cline{2-2} \cline{4-11} 
                          & bicubic ($\times 2$) &  & \cellcolor[HTML]{ECF4FF}16.2 & \cellcolor[HTML]{ECF4FF}4.1& \cellcolor[HTML]{ECF4FF}15.3  & \cellcolor[HTML]{ECF4FF}31.1     &  \cellcolor[HTML]{ECF4FF}18.6  & \cellcolor[HTML]{ECF4FF}4.0               &  \cellcolor[HTML]{ECF4FF}17.8  &  \cellcolor[HTML]{ECF4FF}39.7    \\ \cline{2-2} \cline{4-11} 
                          & bicubic ($\times 4$)    &      & \cellcolor[HTML]{ECF4FF}8.0 & \cellcolor[HTML]{ECF4FF}4.6  & \cellcolor[HTML]{ECF4FF}10.5 & \cellcolor[HTML]{ECF4FF}10.1  & \cellcolor[HTML]{ECF4FF}10.6 & \cellcolor[HTML]{ECF4FF}\textbf{5.7} & \cellcolor[HTML]{ECF4FF}12.8 & \cellcolor[HTML]{ECF4FF}16.7 \\ \cline{2-2} \cline{4-11} 
                          & SRGAN~\cite{SRGan} ($\times 2$)                          &   & \cellcolor[HTML]{ECF4FF}14.8  & \cellcolor[HTML]{ECF4FF}2.6      & \cellcolor[HTML]{ECF4FF}14.3   & \cellcolor[HTML]{ECF4FF}27.9  &  
                          \cellcolor[HTML]{ECF4FF}16.6 & \cellcolor[HTML]{ECF4FF}3.0 & \cellcolor[HTML]{ECF4FF}16.5 & \cellcolor[HTML]{ECF4FF}33.4 \\ \cline{2-2} \cline{4-11} 
                          & DBPN~\cite{SR_DBPN} ($\times 2$)   &  & \cellcolor[HTML]{ECF4FF}15.0  & \cellcolor[HTML]{ECF4FF}3.5 & \cellcolor[HTML]{ECF4FF}14.3 & \cellcolor[HTML]{ECF4FF}27.4 & \cellcolor[HTML]{ECF4FF}16.7 & \cellcolor[HTML]{ECF4FF}3.4 & \cellcolor[HTML]{ECF4FF}16.1 & \cellcolor[HTML]{ECF4FF}32.0  \\ \cline{2-2} \cline{4-11} 
                          & Real-SR~\cite{SR_real_images_1} ($\times 2$)    &  & \cellcolor[HTML]{ECF4FF}14.2  & \cellcolor[HTML]{ECF4FF}2.6   & \cellcolor[HTML]{ECF4FF}12.4     & \cellcolor[HTML]{ECF4FF}29.5     &   \cellcolor[HTML]{ECF4FF}17.3   & \cellcolor[HTML]{ECF4FF}3.6 & \cellcolor[HTML]{ECF4FF}17.0 & \cellcolor[HTML]{ECF4FF}34.1  \\ \cline{2-2} \cline{4-11} 
                          & BSRGAN~\cite{SR_cvpr2021_random} ($\times 2$)        &     & \cellcolor[HTML]{ECF4FF}16.8      & \cellcolor[HTML]{ECF4FF}4.2   & \cellcolor[HTML]{ECF4FF}15.8     & \cellcolor[HTML]{ECF4FF}36.9     &      \cellcolor[HTML]{ECF4FF}20.2  & 
                          \cellcolor[HTML]{ECF4FF}4.8 &  
                          \cellcolor[HTML]{ECF4FF}18.1 &  
                          \cellcolor[HTML]{ECF4FF}40.5           \\ \cline{2-2} \cline{4-11} 
                          & BM3D &   & 10.4      & 0.8      & 6.8      & 27.9 & 10.9 & 0.7 & 8.8 & 35.1\\ \cline{2-2} \cline{4-11} 
                          & Restormer~\cite{Restormer}                           &                                    & 11.4      & 1.2      & 7.2      & 34.8     & 11.9 & 1.4 & 8.9 & 33.4 \\ \cline{2-11} 
                          & \multirow{4}{*}{-}                  & Deg $t$  & 17.6      & 2.3      & 15.4     & 41.9     &  20.9 &   3.1 &    20.3   &  47.6                \\ \cline{3-11} 
                          &                                     & Deg $t$ + N & 17.9   & 2.5      & 15.9     & 42.5     &   21.0 &    3.0    &      20.4    &   48.2      \\ \cline{3-11} 
                          &       & $D_r$ + Detection                            & \cellcolor[HTML]{ECF4FF}17.7      & \cellcolor[HTML]{ECF4FF}\textbf{4.8}      & \cellcolor[HTML]{ECF4FF}15.8     & \cellcolor[HTML]{ECF4FF}41.0     & \cellcolor[HTML]{ECF4FF}21.4  & \cellcolor[HTML]{ECF4FF}5.6 & \cellcolor[HTML]{ECF4FF}19.6 &  \cellcolor[HTML]{ECF4FF}46.3 \\ \cline{3-11} 
                          &                  & \textbf{AERIS}     &   \textbf{18.4}  &  2.7  &  \textbf{16.4}  &     \textbf{42.5}   & \textbf{21.6} &  3.2 &      \textbf{20.4}   &  \textbf{49.0}  \\ \bottomrule
\end{tabular}}

\end{table}

\begin{table}[]
\caption{Comparison with SOTA restoration methods and different training strategies on \textbf{KITTI-d} dataset. Here \textbf{CenterNet} with \textbf{ResNet-18} backbone is adopted. We also show the inference speed (FPS) in the table.}
\label{tab:KITTI-d}
\resizebox{\linewidth}{!}{%
\begin{tabular}{c|cccccc}
\toprule
Methods & -    & bicubic (x2) & bicubic (x4) & SRGAN~\cite{SRGan} (x2) & DBPN~\cite{SR_DBPN} (x2)       & BSRGAN~\cite{SR_cvpr2021_random} (x2) \\ \hline
AP      & 42.2 & 50.6         & 36.5         & 54.3       & 55.6            & 70.8        \\ \hline
FPS     & \textbf{87.4} & 50.0         & 16.2         & 51.1       & 50.6            & 51.4        \\ \midrule
Methods & BM3D & Restormer~\cite{Restormer}    & Deg $t$          & Deg $t$ + N   & $D_r$ +Detection & \textbf{AERIS}     \\ \hline
AP      & 50.9 & 52.6         & 76.0         & 76.6       & 80.0            & \textbf{80.5}       \\ \hline
FPS     & 86.0 & 86.1         & \textbf{87.4}         & \textbf{87.4}       & 43.6            & \textbf{87.4}       \\ \bottomrule
\end{tabular}}
\end{table}

\textbf{Comparison Methods.} 
To evaluate object detectors' robustness under different conditions' degradation, we separately set the multi-degradation evaluation and single-degradation evaluation (see Sec.~\ref{sec:multi-degradation} and Sec.~\ref{sec:single-degradation} for details). We first compare our methods with SOTA image restoration methods:  non-blind SR methods~\cite{SRGan,SR_DBPN,SR_SWIN-IR}, blind SR methods~\cite{SR_real_images_1,SR_cvpr2021_random}, denoise methods~\cite{IRCNN,SR_SWIN-IR,Restormer}. Also in Sec.~\ref{sec:single-degradation}, we separately using different type of restoration methods to handle different type of degradation, for degradation specific comparison. 

On the other hand, we also compare with other training strategies of the network setting, to evaluate their robustness improvement on detection. As it shown in Table~\ref{tab:COCO-d} and Table~\ref{tab:specific}. ``Deg $t$'' corresponds to train the detector with the LR images $t(x)$ random generated from HR images $x$, with the random degradation transformations in Sec.~\ref{sec:degradation}. ``Deg $t$ + N'' means to mix the training data of HR image $x$ and LR images $t(x)$. ``$D_r$ + Detection'' is the structure like~\cite{SR_object_detection} which joint optimize a pre-process SR block and following object detector.

For fairness, all comparison methods adopt the same data augmentation process and same training setting. In the testing stage, all the results are tested on a single RTX 6000 GPU. We compare the speed by reporting the frames per second (FPS) in the experiments, a simple illustration is shown in Table~\ref{tab:KITTI-d}.

\begin{table*}[t]
  \centering
  \caption{Comparison with SOTA restoration methods and SSL methods on COCO~\texttt{val2017} with \textbf{noise}, \textbf{gaussian blur} condition and \textbf{low resolution} condition (down-sampling ratio: \textbf{2} and \textbf{4}). Here \textbf{CenterNet} with \textbf{ResNet-18} backbone was adopted. Higher resolution results are in \textbf{blue} background.}
  \label{tab:specific}
  \begin{subtable}[t]{0.46\linewidth}
  \centering
  \caption{Noise.}
  \label{tab:noise}
  \resizebox{\linewidth}{!}{%
  \begin{tabular}{c | c | c c c c}
    \toprule
    \diagbox{Method}{$\sigma$}  & $(5,50)$ & $15$  & $25$  & $50$ \\
    \midrule
    -    & 22.8 & 26.8 & 23.8 & 15.4  \\ 
    \midrule
    IRCNN~\cite{IRCNN} & 22.6 & 26.8 & 24.2 & 16.8\\ 
    Swin-IR~\cite{SR_SWIN-IR} & 24.2 & 28.0 & 25.6 & 19.3 \\ 
    Restormer~\cite{Restormer}  & 23.8 & 27.6 & 25.1 & 18.9 \\
    \midrule
    Deg $t$ + N & 24.3  & 27.6 & 25.0 & 18.3 \\ 
    $D_r$ + Detection & 24.8  & 28.5 & 25.5 & 19.4 \\ 
    \textbf{AERIS} & \textbf{25.1}  & \textbf{28.7} & \textbf{26.5} & \textbf{20.2} \\ 
    \bottomrule
    \end{tabular}
  }
  
  \end{subtable} \hfill
  \begin{subtable}[t]{0.46\linewidth}
  \centering
  \caption{Blur.}
  \label{tab:blur}
  \resizebox{\linewidth}{!}{%
   \begin{tabular}{c | c | c c c c}
    \toprule
    \diagbox{Method}{$k$} & \texttt{Mix} &  \raisebox{-.3\totalheight}{\includegraphics[width=0.5cm, height=0.5cm]{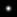}}  & \raisebox{-.3\totalheight}{\includegraphics[width=0.5cm, height=0.5cm]{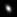}}   & \raisebox{-.3\totalheight}{\includegraphics[width=0.5cm, height=0.5cm]{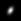}}  \\
    \midrule
    -    & 26.7  & 25.8 & 23.9 & 23.1 \\ 
    \midrule
    EPLL~\cite{EPLL} & 27.8 & 26.8 & 25.4 & 25.2 \\ 
    IRCNN~\cite{IRCNN} & 26.7 & 26.9 & 24.1 & 22.8 \\ 
    
    \midrule
    Deg $t$ + N & 28.8  & 27.5 & 27.6 & 27.8 \\ 
    $D_r$ + Detection & 28.5  & 27.7 & 27.5 & 27.3 \\ 
    \textbf{AERIS} & \textbf{29.3}  & \textbf{28.6} & \textbf{28.0} & \textbf{28.2} \\ 
    \bottomrule
    \end{tabular}
  }
  \end{subtable}
  
    \begin{subtable}[t]{0.46\linewidth}
  \centering
  \caption{Down-sampling (Ratio: \textbf{2}).}
  \label{tab:down2}
  \resizebox{\linewidth}{!}{%
  \begin{tabular}{c | c | c c c c}
    \toprule
    \diagbox{Method}{metric}  & AP & AP$_\text{s}$  & AP$_\text{m}$  & AP$_\text{l}$ \\
    \midrule
    -    & 20.2  & 1.5 & 16.1 & 49.8 \\ 
    \midrule
    SRGAN~\cite{SRGan} ($\times 2$) & \cellcolor[HTML]{ECF4FF}24.0 & \cellcolor[HTML]{ECF4FF}6.2 & \cellcolor[HTML]{ECF4FF}25.4 & \cellcolor[HTML]{ECF4FF}39.8 \\ 
    DBPN~\cite{SR_DBPN} ($\times 2$) & \cellcolor[HTML]{ECF4FF}25.1 & \cellcolor[HTML]{ECF4FF}7.3 & \cellcolor[HTML]{ECF4FF}27.1 & \cellcolor[HTML]{ECF4FF}41.9 \\
    Swin-IR~\cite{SR_SWIN-IR} ($\times 2$)& \cellcolor[HTML]{ECF4FF}25.4 & \cellcolor[HTML]{ECF4FF}7.6 & \cellcolor[HTML]{ECF4FF}27.0 & \cellcolor[HTML]{ECF4FF}42.4 \\ 
    
    \midrule
    $D_r$ + Detection & 26.0  & 8.4 & 26.2 & 46.5 \\ 
    \textbf{AERIS}& 25.4  & 4.6 & 25.8 & \textbf{49.6} \\ 
    \textbf{AERIS} ($\times 2$) & \cellcolor[HTML]{ECF4FF}\textbf{26.8}  & \cellcolor[HTML]{ECF4FF}\textbf{8.6} & \cellcolor[HTML]{ECF4FF}\textbf{28.8} & \cellcolor[HTML]{ECF4FF}45.2 \\ 
    \bottomrule
    \end{tabular}
  }
  
  \end{subtable} \hfill
  \begin{subtable}[t]{0.46\linewidth}
  \centering
  \caption{Down-sampling  (Ratio: \textbf{4}).}
  \label{tab:down4}
  \resizebox{\linewidth}{!}{%
   \begin{tabular}{c | c | c c c c}
    \toprule
    \diagbox{Method}{metric}  & AP & AP$_\text{s}$  & AP$_\text{m}$  & AP$_\text{l}$ \\
    \midrule
    -    & 8.2 & 0.0 & 3.2 & 33.1 \\ 
    \midrule
    DBPN~\cite{SR_DBPN} ($\times 2$) & \cellcolor[HTML]{ECF4FF}14.8 & \cellcolor[HTML]{ECF4FF}1.0 & \cellcolor[HTML]{ECF4FF}9.9 & \cellcolor[HTML]{ECF4FF}39.7 \\ 
    DBPN~\cite{SR_DBPN} ($\times 4$) & \cellcolor[HTML]{ECF4FF}12.2 & \cellcolor[HTML]{ECF4FF}1.7 & \cellcolor[HTML]{ECF4FF}11.7 & \cellcolor[HTML]{ECF4FF}23.4 \\ 
    Swin-IR~\cite{SR_SWIN-IR} ($\times 2$) & \cellcolor[HTML]{ECF4FF}15.2 & \cellcolor[HTML]{ECF4FF}1.1 & \cellcolor[HTML]{ECF4FF}10.1 & \cellcolor[HTML]{ECF4FF}39.9 \\ 
    Swin-IR~\cite{SR_SWIN-IR} ($\times 4$) & \cellcolor[HTML]{ECF4FF}12.8 & \cellcolor[HTML]{ECF4FF}1.8 & \cellcolor[HTML]{ECF4FF}12.2 & \cellcolor[HTML]{ECF4FF}23.4 \\ 
    \midrule
    $D_r$ + Detection & \cellcolor[HTML]{ECF4FF}15.1  & \cellcolor[HTML]{ECF4FF}1.8 & \cellcolor[HTML]{ECF4FF}12.7 & \cellcolor[HTML]{ECF4FF}40.1 \\ 
    \textbf{AERIS} & 13.0  & 0.8 & 10.2 & \textbf{42.6} \\
    \textbf{AERIS} ($\times 2$) & \cellcolor[HTML]{ECF4FF}\textbf{15.8}  & \cellcolor[HTML]{ECF4FF}\textbf{2.0} & \cellcolor[HTML]{ECF4FF}\textbf{13.2} & \cellcolor[HTML]{ECF4FF}40.9 \\ 
    \bottomrule
    \end{tabular}
  }
  \end{subtable}
\end{table*}

\subsection{Multi-Degradation Evaluation}
\label{sec:multi-degradation}

To evaluate object detectors' robustness under diverse degradation of real-world images. Different from previous works~\cite{ICLR_2019_robustness,robust_segmentation_1,noise_4_vision,CVPR_blur_detection} that only consider single degradation type at a time. Following the practical degradation model~\cite{Degradation_model,Degradation_model_1} in Eq.\ref{eq:degradation_model}, we design the experiments on images with multiply degradation conditions and down-scale ratios, to verify detection robustness in real-world diverse condition. 
We generate \textbf{COCO-d} dataset from original COCO~\texttt{val2017} dataset and \textbf{KITTI-d} dataset from original KITTI~\texttt{val} dataset. We give per-image a random blur kernel (isotropic Gaussian kernel $k_{iso}$, anisotropic Gaussian kernel $k_{aniso}$) and random noise level (AWGN noise with variance $\sigma \sim U(0, 25/255)$). As for the resolutions, we down-sample per image in COCO~\texttt{val2017} with a random rate $s \sim U(1.0, 4.0)$.

The experimental results are shown in Table~\ref{tab:COCO-d} and Table~\ref{tab:KITTI-d}, we add the up-scale ratio ($\times 2$, $\times 4$) after name of SR methods. We first give the detection results on original COCO~\texttt{val2017} and \textbf{COCO-d} dataset. The object detector is easily affected and the performance suffers  a large decrease on the multi-degradation condition. Table~\ref{tab:COCO-d} also verifies up-scale higher resolution (either by interpolate or SR pre-process) 
improves the detection performance on small objects, but has a negative impact on the middle and large objects. Restoration methods would also invalid if degradation types and down-sampling scales are diverse, among several restoration methods, real-world SR method BSRGAN~\cite{SR_cvpr2021_random} could get satisfactory results. Our AERIS model could get best performance in most of metrics, even with a lower input resolution, but the one limitation is that AERIS could not get best performance on small object.

\subsection{Degradation Specific Evaluation}
\label{sec:single-degradation}

To further understand the advantages of AERIS, we design the experiments on single degradation conditions. We separately make experiments on noise, gaussian blur and low-resolution conditions. For noise and blur condition, in training stage of three SSL methods, we generate $t(x)$ from $x$ with noise $n$ (variance  $\sigma \sim U(0, 50/255)$) and blur kernel $k$ (same as Sec.~\ref{sec:degradation}). As for low resolution condition, we generate $t(x)$ from $x$ with down-sampling ratio $s$. Here we discuss three conditions as follow:

\textbf{Performance \textit{w.r.t} Noise $\&$ Blur.}
For noise's affect on object detection, we process COCO~\texttt{val2017} with random Gaussian noise $n \sim N(0, \sigma)$, we first random choose variance $\sigma$ from uniform distribution $U(5/255, 50/255)$ for mix noise level evaluation. Then we take three different noise level: $\sigma=15$, $\sigma=25$ and $\sigma=50$ for specific evaluation. We compare AERIS with SOTA denoise methods IRCNN~\cite{IRCNN}, Swin-IR~\cite{SR_SWIN-IR} and Restormer~\cite{Restormer} and also compare with other training strategies. We report the average precision (AP) in Table~\ref{tab:noise}.

For blur's affect on object detection, we first process COCO~\texttt{val2017} with random isotropic Gaussian kernel $k_{iso}$ and anisotropic Gaussian kernel $k_{aniso}$ (probability both 0.5) as \texttt{Mix} evaluation. Then we specifically choose three degradation kernels for specific evaluation (see Table~\ref{tab:blur}). We compare with Gaussian deblur methods EPLL~\cite{EPLL} and IRCNN~\cite{IRCNN} and other two other training strategies, then report AP value in Table~\ref{tab:blur}. Our AERIS gains \textbf{best} performance under various noise and blur conditions, among image restoration methods and other training strategies.

\textbf{Performance \textit{w.r.t} Low Resolution.}
To evaluate low resolution's affect on detection task, we down-scale original COCO~\texttt{val2017} with a fixed down-sampling ratio $s$. Here we set the down-sampling ratio to $2$ and $4$, and then compare with SOTA SISR methods SRGAN~\cite{SRGan}, DBPN~\cite{SR_DBPN}, Swin-IR~\cite{SR_SWIN-IR} and pre-stage method ``$D_r$ + Detection''. We report the total AP and different level objects' detection performance (AP$_\text{s}$, AP$_\text{m}$, AP$_\text{l}$) in Table~\ref{tab:down2} and Table~\ref{tab:down4}. We also make an additional experiments to up-scale input images with ratio $2$ and then send into AERIS model as \textbf{AERIS ($\times 2$)}, for same resolution comparison with SR methods and pre-upsampling method.

% \usepackage{multirow}
\subsection{Ablation Study}

% \begin{table}[t]
% \captionsetup{font={scriptsize}}
% \caption{Ablation results of CenterNet architecture on MS COCO and KITTI.}

% \centering
% \begin{adjustbox}{max width= 1 \linewidth}
% \begin{tabular}{ccccc|c|c}
% \toprule
% CenterNet & + Deg $t$ & + $D_t$ & \begin{tabular}[c]{@{}c@{}}+ feature\\ connect\end{tabular} & + ARRD & \begin{tabular}[c]{@{}c@{}}mAP\\ (COCO-d)\end{tabular} & \begin{tabular}[c]{@{}c@{}}mAP\\ (KITTI-d)\end{tabular} \\ \hline
%  $\surd$  &    &    &   &      & 14.5 &  46.2  \\ \hline
%  $\surd$  & $\surd$ &    &   &      & 16.8 (+2.3) & 74.5 (+28.3) \\ \hline
%  $\surd$  & $\surd$  & $\surd$  &  &     & 17.0 (+2.5)& 74.9 (+28.7)\\ \hline
%  $\surd$  & $\surd$  &    &  $\surd$  &  & 17.5 (+3.0)& 76.0 (+29.8)\\ \hline
%  $\surd$  & $\surd$  &    &    & $\surd$ & 17.5 (+3.0)& 75.8 (+29.6)\\ \hline
%  $\surd$  & $\surd$  &    & $\surd$ & $\surd$ & 18.1 (+3.6)& 79.8 (+33.6)\\ \hline
%   $\surd$  & $\surd$  &  $\surd$  & $\surd$ & $\surd$ & 18.2 (+3.7)& 80.5 (+34.3)\\ \bottomrule
% \end{tabular}
% \end{adjustbox}
% \label{fig:Ablation}

% \end{table}
\begin{table}[t]
% \captionsetup{font={scriptsize}}
\caption{Ablation results of CenterNet on \textbf{COCO-d} dataset.}
\centering
\begin{adjustbox}{max width= 1 \linewidth}
\begin{tabular}{ccccccc}
\toprule
CenterNet & +Deg $t$ & \begin{tabular}[c]{@{}l@{}}+feature\\ connect\end{tabular} & \begin{tabular}[c]{@{}c@{}}+ARRD\\ loc 1\end{tabular} & \begin{tabular}[c]{@{}c@{}}+ARRD\\ loc 2\end{tabular} & \begin{tabular}[c]{@{}c@{}}+ARRD\\ loc 3\end{tabular} & \begin{tabular}[c]{@{}c@{}}mAP\\ (COCO-d)\end{tabular} \\ \hline
   $\surd$  &         &        &       &       &     & 14.5      \\ \hline
    $\surd$  &  $\surd$  &       &        &       &     & 16.8 (+2.3)   \\ \hline
    $\surd$  &  $\surd$  &    $\surd$   &        &       &     & 17.5 (+3.0)     \\ \hline
    $\surd$ &   $\surd$ &        &        &  $\surd$  &    & 17.7 (+3.2)     \\ \hline
    $\surd$ & $\surd$  &  $\surd$  & $\surd$   &    &       & 18.2 (+3.7)     \\ \hline
    $\surd$ &  $\surd$  &  $\surd$ &      & $\surd$  &       & 18.4 (+3.9)   \\ \hline
    $\surd$ &  $\surd$  &  $\surd$  &       &       &  $\surd$  & 17.9 (+3.4)      \\ \bottomrule
\end{tabular}
\end{adjustbox}
\label{tab:Ablation}
\end{table}

To evaluate the location of ARRD on CenterNet~\cite{centernet}, we separately make the ablation study to evaluate each part's effectiveness, as shown in Table~\ref{tab:Ablation}, ``+Deg $t$'' refers to adding the degradation transformation in Sec.~\ref{sec:degradation}.  ``+feature connect'' means the feature connection process in Fig.~\ref{fig:detail_structure}. ``+ARRD''  adds the decoder $D_r$ upon the network structure. We also evaluate adding ARRD on different location of the up-sampling blocks, ``loc 1'' means the shallow up-samling layer, ``loc 2'' means middle up-sampling layer and ``loc 3'' means the final up-sampling layer (``loc 3'' also connect with the detection decoder $D_o$), as our finding, to implement ARRD on the middle layer could get best result.

\section{Conclusion}
In this paper, we propose a novel self-supervised framework, AERIS, to handle object detection for arbitary degraded low resolution images. To capture the dynamics of feature representations under diverse resolution and degradation conditions, we propose a degradation equivariant representation that is generic and could be implemented on popular detection architectures. To further combine the strength of the existing progress on super resolution (SR), we  also introduce an arbitrary-resolution restoration decoder that supervises the latent representation to preserve the visual structure. The extensive experiments demonstrate that  our AERIS achieves SOTA results on  two mainstream public datasets among different degradation conditions (resolution, noise and blur).

\section{Acknowledgement}
This work was supported by JST Moonshot R\&D Grant Number JPMJMS2011 and JST ACT-X Grant Number JPMJAX190D, Japan.

\bibliographystyle{splncs04}
\bibliography{egbib}

\newpage
\appendix

\renewcommand\thesection{\Alph{section}}
\renewcommand\thetable{\Alph{section}\arabic{table}} 
\renewcommand\thefigure{\Alph{section}\arabic{figure}}

\section{Results on FPN structure.}

We evaluate AERIS's effectiveness on CenterNet~\cite{centernet} in the main text, where the CenterNet structure only take one level feature map for detection decoder $D_o$, also single-level feature structure could better illustrate how our
self-supervised signal directly enhances the performance. Further more, more recent object detectors take FPN~\cite{detection_FPN} and FPN series structures for detection decoding, take advantage of the multi-branch design, different level feature maps would outputs multi object detection results.

We evaluate our method on
RetinaNet-FPN with Swin-T backbone for the multi-level
scenario, name as AERIS-FPN. The experiments are done on the \textbf{COCO-d} dataset, we trained the model for 24 epochs with SGD optimizer, the initial learning rate is 0.01 and decay to one-tenth at 16 and 22 epoch, data augmentation is same as \texttt{mmdetection}~\cite{mmdetection}, other setting is same as Table 1 in main text. For evaluating different places of FPN to add the ARRD decoder $D_r$, Different level output features
of FPN (1, 2, 3, 4, 5) are used for the reconstruction (1 is
highest resolution feature map, 2 is second high resolution
feature map and so on), we implement ARRD decoder on single level feature map (1) and mixer of multi-level feature map (1,2) and (1,2,3) and so on. The experiment results has been shown in Table~\ref{fig:FPN}, we are
glad our framework achieves a consistent improvement also
on multi-level scenario. This strength our belief that the
proposed strategy could leverage the recent advances in vision architectures. 

\begin{table}[]
% \captionsetup{font={scriptsize}}
\caption{Experiments on RetinaNet-FPN structure on \textbf{COCO-d} dataset, we compare the ARRD with different level inputs from FPN.}

\centering
\begin{adjustbox}{max width= 1 \linewidth}
\begin{tabular}{c|c|c|c|c|c}
\toprule
Metrics & w/o ARRD & ARRD (1,2,3,4) & ARRD (1,2,3) & ARRD (1, 2) & ARRD (1) \\ \hline
AP  & 21.5     & \textbf{22.1}           & 22.0         & 21.7        & 22.0     \\ \hline
AP$_\text{s}$ & 2.5      & 2.8            & 2.8          & 2.8         & \textbf{3.0}      \\ \hline
AP$_\text{m}$ & 19.0     & \textbf{19.7 }          & 19.4         & 19.3        & 19.4     \\ \hline
AP$_\text{l}$ & 46.0     & \textbf{47.1}           & 46.5         & 46.3        & 46.8     \\ \bottomrule
\end{tabular}
\end{adjustbox}

\label{fig:FPN}
\end{table}

\section{Restoration Network Fine-tune.}
\setcounter{table}{0}

\begin{table}[]
\centering
\caption{Experiments compare with fine-tuned restoration methods on \textbf{COCO-d} dataset.}
\begin{adjustbox}{max width= 1 \linewidth}
\begin{tabular}{ccccccc}
\hline
       & SRGAN~\cite{SRGan} & DBPN~\cite{SR_DBPN}  & Real-SR~\cite{SR_real_images_1}  & BSRGAN~\cite{SR_cvpr2021_random} & Restormer~\cite{Restormer} & Ours                  \\ \hline
w/o FT & 14.8 & 15.0 & 14.2 & 16.8   & 11.4      & \multirow{2}{*}{18.4} \\ \cline{1-6}
w FT   & 15.5 & 16.4 & 15.4 & 16.9   & 12.3     &                       \\ \hline
\end{tabular}
\end{adjustbox}
\label{fig:FT}
\end{table}

Avoiding fixed degradation parameter is actually the unique advantage of our method, since our AERIS framework directly finds intrinsic equivariant representation against various resolutions and degradation. To compare with existing image restoration methods as fair as possible,
 we further fine-tune the pre-train image restoration network~\cite{SRGan,SR_DBPN,SR_real_images_1,SR_cvpr2021_random,Restormer} on the multi-degradation settings, same as the dataset generation of \textbf{COCO-d} except the resolution factor $s$, since the up-sampling resolution is often fixed in super-resolution network. We then make experiments on \textbf{COCO-d} dataset and results are shown in Table.~\ref{fig:FT}. The fine-tune process would further improve the  image restoration networks' performance, and our AERIS also gain best performance among different restoration methods.

\section{Different Scale in Training.}
\setcounter{table}{0}

In AERIS training, we choose the down-sampling ratio $s$ from a uniform distribution $s \sim (1, 4)$, for ablation study we further make the experiments with the different down-sampling scale range in training stage, the experimental results on \textbf{COCO-d} dataset are shown in Table~\ref{fig:Scale}, $s = 1$ means keep the original resolution and $s \sim (1, x)$ means to choose $s$ from  an uniform distribution $U (1, x)$.

\begin{table}[]
\centering
\caption{Ablation study about different training scale on \textbf{COCO-d} dataset.}
\begin{adjustbox}{max width= 1 \linewidth}
\begin{tabular}{ccccc}
\toprule
s  & $\sim$(1, 4) & $\sim$(1, 3) & $\sim$(1, 2) & 1   \\ \hline
AP & 18.4         & \textbf{18.5}        & 17.8         & 17.6 \\ \hline
AP$_\text{s}$ & \textbf{2.7}          & 2.5          & 2.3          & 2.0  \\ \hline
 AP$_\text{m}$  & \textbf{16.4}         & 16.0         & 15.0         & 14.8 \\ \hline
 AP$_\text{l}$  & 42.5         & \textbf{44.0}        & 43.2         & 41.9 \\ \bottomrule
\end{tabular}
\end{adjustbox}
\label{fig:Scale}
\end{table}

\end{document}